\documentclass[conference]{IEEEtran}

\usepackage{amsmath,amssymb,amsfonts}
\usepackage{algorithmic}
\usepackage{graphicx}
\usepackage{textcomp}
\usepackage{xcolor}
\usepackage{verbatim}
\usepackage{subcaption}
\usepackage{cite}

\def\BibTeX{{\rm B\kern-.05em{\sc i\kern-.025em b}\kern-.08em
    T\kern-.1667em\lower.7ex\hbox{E}\kern-.125emX}}

\newcommand{\R}{\mathbb{R}}
\newcommand{\sat}{\mathrm{sat}}
\newcommand{\diag}{\mathrm{diag}}

\newcommand{\Prb}{\mathbb{P}}

\begin{document}

\title{Modeling and Validation of Quality of Control for Edge-Offloaded Collaborative Navigation}
\author{
    \IEEEauthorblockN{
        Neelabhro Roy\IEEEauthorrefmark{2},  
        Mikael Hammarling\IEEEauthorrefmark{2},
        Victor Nan Fernandez-Ayala\IEEEauthorrefmark{2},
        Gourav Prateek Sharma\IEEEauthorrefmark{1},\\
        Mani H. Dhullipalla\IEEEauthorrefmark{2}, 
        Dimos V. Dimarogonas\IEEEauthorrefmark{2} and
        James Gross\IEEEauthorrefmark{2}
    } 
        \IEEEauthorblockA{
        \IEEEauthorrefmark{2}%\textit{School of Electrical Engineering \& Computer Science} \\
        \text{KTH Royal Institute of Technology, Stockholm, Sweden }}

    \IEEEauthorblockA{
        \IEEEauthorrefmark{1}%\textit{Department of Radio Access Propagation} \\
        \text{National Institute of Technology, Kurukshetra, India  }\\
        Email: \{nroy, mhammarl, vnfa, manihd, dimos, jamesgr\}@kth.se, gourav.sharma@nitkkr.ac.in\\}
        %Stockholm, Sweden \\
    
}

    %\IEEEauthorblockA{
    %    \IEEEauthorrefmark{1}Ericsson AB, \{jiayi.tan, rohit.chandra, jason.t.chen\}@ericsson.com \\
    %    \IEEEauthorrefmark{2}KTH Royal Institute of Technology, \{nroy, jamesgr\}@kth.se\\
    %    Stockholm, Sweden 
    %}

\maketitle

\begin{abstract}
%The control of multiple collaborative robots in complex environments becomes substantially harder when the motion and navigation stack depends on a wireless network whose delay and reliability vary stochastically over space and time. Such variations affect navigation, trajectory tracking, dynamic collision avoidance, and also the energy spent to achieve collaborative objectives such as consensus, formation control, target tracking with collision avoidance, etc. In this paper, for a navigation setup with dynamic collision avoidance, we address this challenge through a quality of control (QoC) framework that (i) models end-to-end network effects on closed-loop system performance, (ii) validates QoC simulations experimentally against measured multi-robot runs over a private 5G testbed. We realize this by fundamentally expanding the QoC definition in \cite{qoc_initial}, to be applicable for practical robotic setups by considering a unicycle robotic model.
Collaborative control in complex environments is severely challenged by stochastic wireless delay and reliability variations, which can degrade navigation, tracking, and collision avoidance. These network-induced uncertainties complicate the maintenance of energy efficiency during collaborative tasks, and can potentially lead to over-provisioning of resources. In this paper, for a navigation setup with dynamic collision avoidance, we address this challenge by expanding the quality of control (QoC) framework from prior works to practical robotic models. Our approach (i) models end-to-end network effects on closed-loop performance, (ii)  systematically explores the impact of various control parameters dictating robotic motion on network latency-reliability (iii) validates these models through experiments on a private 5G testbed across varying delay, reliability and control configurations. Our analysis indicates the optimal control-communication co-design operating regimes for practical robots and also compares the QoC performance of standard ROS~2 quality of service (QoS) policies under real-world conditions and showing how RELIABLE QoS offers 51.5\% better QoC than BEST-EFFORT under certain experimental settings.
%, further incorporating velocity caps and information loop completion probabilities for collaborative robotic systems. 
%The proposed simulation framework is then validated with extensive experiments spanning different delays, reliability and various control parameters under real 5G network settings, also offering insights on how built-in ROS functions fare against each other in terms of QoC.
%Finally, we use QoC thresholds learned from experiments to switch between ROS~2 QoS reliability profiles (BEST\_EFFORT vs RELIABLE) so that the system trades off latency and delivery guarantees in a principled manner, optimizing for energy efficiency under adverse network conditions.
%map live network measurements into predicted QoC during execution, and integrate this into a QoC- and network-aware costmap and MPPI-based local control. 
\end{abstract}
% =========================================================
\section{Introduction}
% =========================================================
With the emergence of cloud and edge computing promising low latency and ultra reliable wireless communications, several works have investigated offloading diverse workloads, ranging from perception to motion planning for industrial automated guided vehicles (AGVs) and collaborative robots to edge/cloud systems \cite{baxi2022towards, codesign1, qoc_initial}. Such architectures enable advanced capabilities like external sensing, centralized mapping, and intensive compute offloading, thereby reducing the onboard compute burden while broadening robotic capabilities, important for flexible automation and production \cite{qoc_initial,ros_distributed, zhihao}. However, the shift from rigid, wired industrial setups to adaptive wireless solutions introduces significant stochasticity in the form of delay, jitter, and packet loss \cite{qoc_initial, QoC_Assessment}. These uncertainties directly affect sensing, navigation, and, most critically, the stability of control loops in coordination or collaboration \cite{qoc_initial, roy2026qoc_codesign}.

Effective collaborative task execution remains a challenging problem and often relies on a centralized task planner in an edge server responsible for mapping, path planning, and navigation \cite{mohanti2023norm}. A fundamental limitation and design parameter in these deployments is the robotic energy budget, due to battery constraints, which further warrants efficient path planning to prevent energy waste. Moreover, managing energy expenditure is not merely a hardware concern originating from battery issues, but a crucial constraint for wireless collaborative robotics (WCRs) \cite{qoc_initial} in general.
As control offloading to the edge increases, managing communication-induced uncertainty becomes vital for state and energy regulation \cite{qoc_initial}. While energy-centric studies \cite{collab4g, carabin2017review} optimize search-and-rescue efficiency assuming the control loop is less time-sensitive and robust to typical 5G latencies, they overlook how stochastic delay and reliability shape closed-loop coordination. Recent distributed evaluations, such as the robot operating system (ROS 2)-based 5G image-processing stack in \cite{ros_distributed}, highlight the complex factors influencing reaction times but rely on simple safety timeouts to mitigate jitter. Existing literature \cite{ros_distributed, cleave, latency_min} largely analyzes isolated parameters like sampling rate or latency for individual robots. Consequently, there is still a lack of a comprehensive abstraction that jointly accounts for communication and control parameters to quantify collaborative system performance and energy expenditure.
Quality-of-Control (QoC)-based metrics and abstractions have been studied in recent works \cite{qoc_initial, QoC_Assessment, roy2026qoc_codesign} linking wireless parameters to robotic behaviour. By unifying energy and control expenditure into a single objective, QoC further enables energy gains by optimizing the delay-reliability trade-off as shown in \cite{roy2026qoc_codesign}. However, this approach was limited to simulations only, considered simplified consensus dynamics and lacked experimental validation in \cite{qoc_initial, roy2026qoc_codesign}. Crucially, it overlooked non-holonomic AGV dynamics, where the non-linear coupling of orientation and translation makes tracking more sensitive to network stochasticity than simpler linear, holonomic models \cite{AGV_model}. 
To address these gaps, this work develops a system model tailored for practical robotic motion and validates its effectiveness through both extensive simulations and practical 5G testbed experiments. Our main contributions are summarized as follows:

\begin{itemize}
    \item {Non-Holonomic System Modeling:} We generalize the frameworks in \cite{qoc_initial, roy2026qoc_codesign} by incorporating non-holonomic kinematic constraints. Unlike previously considred holonomic systems, which can translate instantaneously in any direction, our proposed model accounts for motion restricted by the robot's current heading. This enables the characterization of network stochasticity on coupled translational and rotational AGV dynamics, together.
    \item {QoC-Based Analysis:} We propose and derive a QoC abstraction for 5G-Edge offloaded navigation featuring dynamic collision avoidance, quantifying the specific impact of various control parameters and network-induced impairments on system performance, towards co-design.
    \item {Experimental Validation:} We validate our simulation framework using a private 5G testbed \cite{expeca} and the Nav2 architecture. The experimental results confirm the framework's utility, further demonstrating that ROS~2 RELIABLE QoS yields a 51.5\% higher QoC compared to BEST-EFFORT under real-world settings.
\end{itemize}
% =========================================================
\section{System Model}
\label{system_model}
\begin{figure}[t]
    \centering
    \includegraphics[width=0.99\linewidth]{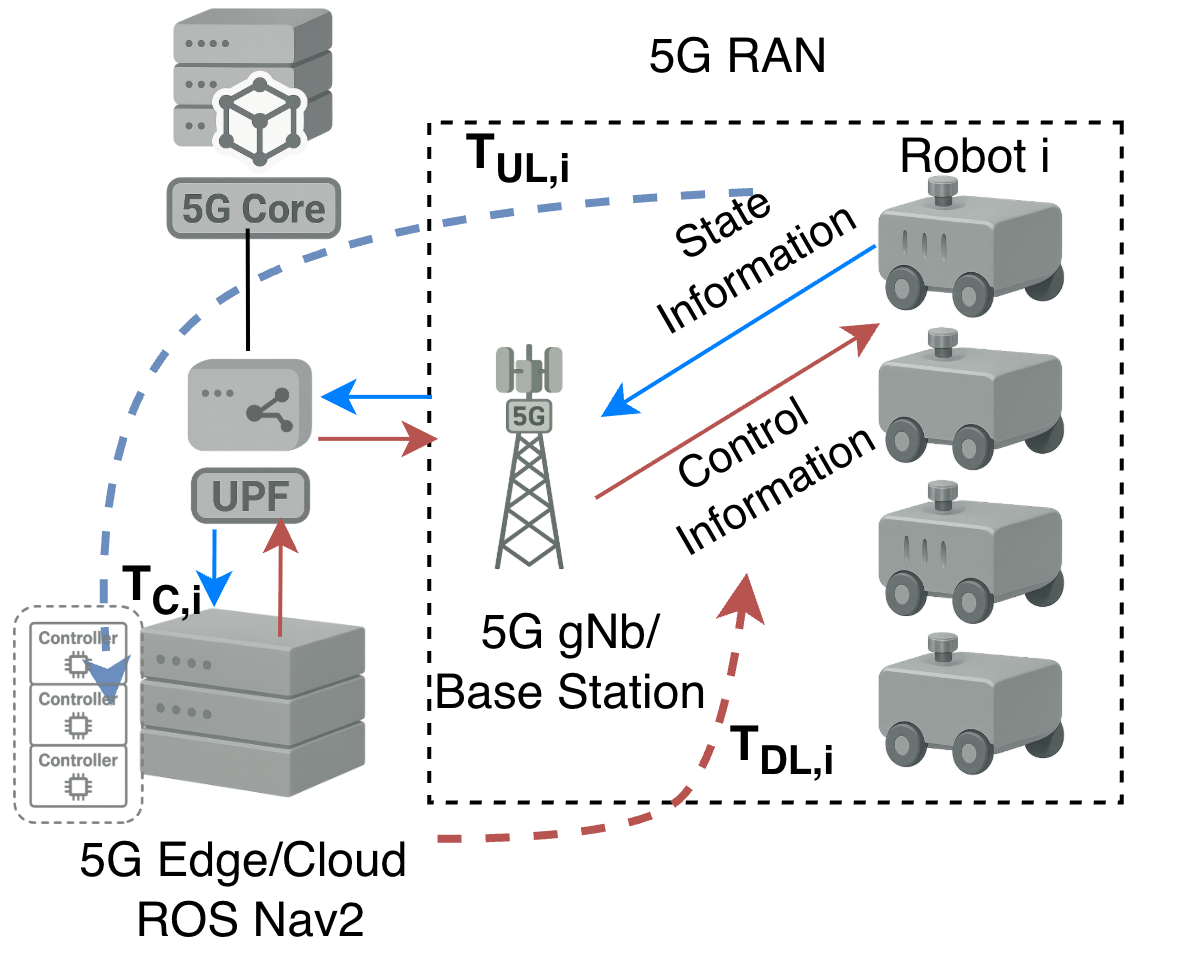}
    \caption{Overview of a WCR setup with control offloaded to a centralized server running Nav2. $T_{\mathrm{UL},i}$, $T_{\mathrm{C},i}$ and $T_{\mathrm{DL},i}$ have been defined in Section \ref{communication_model}. }\label{fig:sys_model}
\end{figure}

% =========================================================
In this work, we model a WCR system, illustrated in Fig.~\ref{fig:sys_model}, which comprises robots collaborating over a shared 5G wireless network. This section discusses key system aspects, including the robot dynamics and navigation, aspects of the 5G network forming the underlying communication channel, and finally, the problem statement.

\subsection{Overview and Probem Statement}
%Our study considers a multi-robot navigation system where $N$ AGVs (Turtlebot 3\footnote{https://www.turtlebot.com/} in experiments) perform a navigation task (e.g., arriving at a specified destination) with dynamic collision avoidance. The controllers for these AGVs run on an edge compute system for which control commands are periodically transmitted over a private 5G network \cite{expeca}. The end-to-end information loop is stochastic, induces both delay and reliability variations, modeled explicitly in our framework.
%We consider $N$ AGVs (TurtleBot 3 in the subsequent experiments\footnote{https://www.turtlebot.com/}) performing a navigation task (e.g., arriving at a specified common destination) with dynamic collision avoidance. This setup utilizes an edge-based architecture where computation is offloaded to a 5G controller, facilitating periodic message exchanges over a private 5G network \cite{expeca}. Central to our approach is the explicit modeling of the stochastic delay and reliability variations characterizing this end-to-end information loop, ensuring the framework captures the inherent uncertainties of the wireless communication medium, since the robots do not directly communicate with each other and instead collaborate with each other via the Edge controller.
Offloading control to the 5G edge enables advanced collaboration but introduces stochastic uncertainties that can compromise stability and system performance. Existing frameworks often overlook the non-linear coupling of orientation and translation in practical AGVs, leaving a gap between theoretical models and real-world navigation. This paper addresses this by modeling and validating a QoC-based abstraction for non-holonomic motion to identify optimal control-communication regimes for co-design. We consider $N$ AGVs (TurtleBot 3\footnote{https://www.turtlebot.com/}) performing navigation with dynamic collision avoidance via an edge-based controller. Because the robots collaborate exclusively through this centralized offloading rather than peer-to-peer links, our framework explicitly models the stochastic delay and reliability of the periodic, end-to-end information loop over a private 5G network \cite{expeca}.

\begin{figure}[t]
    \centering
    \includegraphics[width=0.99\linewidth]{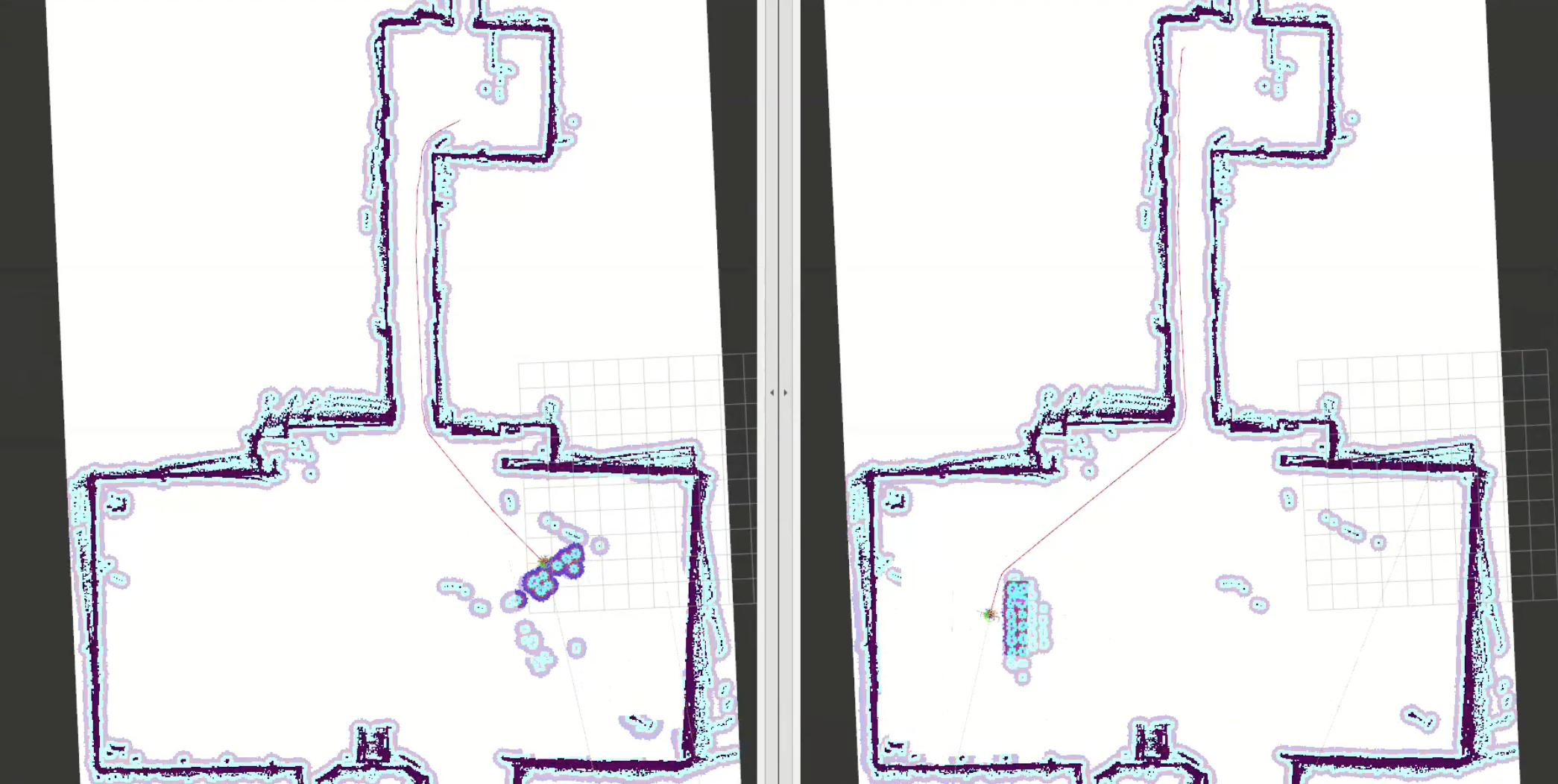}
    \caption{Illustration of the costmap utilized by two different robots for Nav2-based navigation.}\label{costmap}
\end{figure}

\subsection{Robot Dynamics and Control}
\label{dynamics}
\subsubsection{Modeling robot dynamics}
We extend the simulation framework discussed in \cite{qoc_initial, roy2026qoc_codesign} towards modeling of AGV motion observed in practice. We consider planar and angular motion of robots $i\in\{1,\dots,N\}$ under an AGV kinematic model \cite{AGV_model} with state and commanded inputs:
\begin{equation}
\mathbf{x}_i(t) \triangleq [\,x_i(t)\;\;y_i(t)\;\;\theta_i(t)\,]^\top \in \R^3,
\end{equation}

\begin{equation}
\mathbf{u}_i(t)\triangleq [\,\nu_i(t)\;\;\omega_i(t)\,]^\top\in\R^2,
\end{equation}
where $(x_i,y_i)$ is the planar position, $\theta_i$ is the heading angle, and $(\nu_i,\omega_i)$ denote the translational and the rotational speeds, respectively. 
The state evolution \cite{AGV_model} is given by: 
\begin{equation}
\dot{\mathbf{x}}_i(t)=J(\theta_i(t))\,\mathbf{u}_i(t),
\qquad
J(\theta)\triangleq
\begin{bmatrix}
\cos\theta & 0\\
\sin\theta & 0\\
0 & 1
\end{bmatrix},
\label{jacobian}
\end{equation}
i.e., $\dot{x}_i=\nu_i\cos\theta_i$, $\dot{y}_i=\nu_i\sin\theta_i$, and $\dot{\theta}_i=\omega_i$.

Let $\mathbf{p}_i(t)\triangleq[\,x_i(t)\;\;y_i(t)\,]^\top$.
At discrete sampling instants $t_p = t_0 + p\tau$, where $t_0$ is the initial time, $p$ is the sampling index, and $\tau$ is the fixed sampling period, the edge-based controller (Fig. \ref{fig:sys_model}) computes the control command, applied via a zero-order hold (ZOH), such that $\mathbf{x}_i(t)$ evolves as:
\begin{equation} 
\label{eq:robotdyn ZOH} \dot{\mathbf{x}}_i(t)=J(\theta_i(t)),\mathbf{u}_i(t_p), \quad t\in[t_p,t_{p+1}), 
\end{equation}
where $J(\theta_i)$ is the Jacobian mapping commanded velocities to the global frame, as shown in Eq.\eqref{jacobian}.

\subsubsection{Control commands}
In this paper, the objective of the robots in the system is to arrive at their pre-specified destinations by successfully avoiding collisions. To facilitate this, prior to designing appropriate control commands $u_i(t_p)$ in Eq.\eqref{eq:robotdyn ZOH}, we define the following variables:

At sampling instants $t_p$, the edge controller computes the desired planar velocity $\mathbf{v}_i^{\mathrm{des}}(t_p)$ for each robot $i$ moving towards the target coordinate $\mathbf{p}^\star \in \mathbb{R}^2$ using estimated positions $\hat{\mathbf{p}}_i(t_p)$:
\begin{equation}
\mathbf{v}_i^{\mathrm{des}}(t_p)
\triangleq
-\kappa\big(\hat{\mathbf{p}}_i(t_p)-\mathbf{p}^\star\big)
+
\sum_{j\neq i}\mathbf{v}^{\mathrm{rep}}_{ij}(t_p),
\label{eq:vdes_goal_rep}
\end{equation}
where $\kappa>0$ is the goal gain capturing the strength of the closed-loop response, also absorbing unmodeled nonlinearities\footnote{Velocity commands are generated by the Nav2 stack (A*/MPPI) running on the edge. In simulation, we can use the analytical feedback law in this section as a surrogate controller to isolate QoC sensitivity to delay/reliability while matching the measured timing and loop-completion statistics.}, and $\mathbf{v}^{\mathrm{rep}}_{ij}$ is a repulsive collision-avoidance term, tackling potential collisions between robots $i$ and $j$. Note that Eq.\eqref{eq:vdes_goal_rep} is inspired from robot control for obstacle avoidance using artificial potential fields (APFs) \cite{apf}. Let $d_{ij}(t_p)\triangleq \|\hat{\mathbf{p}}_i(t_p)-\hat{\mathbf{p}}_j(t_p)\|_2$ be the estimated inter-robot distance and let $d_{\mathrm{safe}}^{\mathrm{eff}}$ be the  effective safety radius that expands with information staleness:
\begin{equation}
d_{\mathrm{safe}}^{\mathrm{eff}}(t_p)
\triangleq
d_{\mathrm{safe}} + c_{\mathrm{miss}} \max\Big(0,\max(a_i(t_p),a_j(t_p))-\Delta\Big),
\label{eq:dsafe_eff}
\end{equation}
where $d_{\mathrm{safe}}>0$ is the nominal safety distance, $c_{\mathrm{miss}}\ge 0$ is a penalty coefficient, and $\Delta$ is the loop completion deadline. The age variable $a_i(t)$ captures the time since the last successful control update (Eq.\eqref{loop_completion}). This keeps robots from violating a minimum distance \cite{apf}, and becomes more conservative under packet drops via Eq.\eqref{eq:dsafe_eff}. For a repulsion gain $\eta>0$:
%, the avoidance term is defined as:
\begin{equation}
\mathbf{v}^{\mathrm{rep}}_{ij}(t_p)\triangleq
\begin{cases}
\eta\Big(\frac{1}{d_{ij}}-\frac{1}{d_{\mathrm{safe}}^{\mathrm{eff}}}\Big)
\frac{\hat{\mathbf{p}}_i(t_p)-\hat{\mathbf{p}}_j(t_p)}{d_{ij}^3},
& d_{ij}<d_{\mathrm{safe}}^{\mathrm{eff}},\\[4pt]
0, & \text{otherwise.}
\end{cases}
\label{eq:repulsion}
\end{equation}

To accommodate AGV kinematics, $\mathbf{v}_i^{\mathrm{des}}$ is mapped to translational and rotational speeds $(v_i, \omega_i)$. The desired heading angle $\theta_i^{\mathrm{des}}(t_p)$ and the orientation error $e_{\theta,i}(t_p)$ relative to the estimated heading $\hat{\theta}_i(t_p)$ are given by:
\begin{equation}
\theta_i^{\mathrm{des}}(t_p)\triangleq
\mathrm{atan2}\!\big(v^{\mathrm{des}}_{y,i}(t_p),\,v^{\mathrm{des}}_{x,i}(t_p)\big)  
\label{heading}
\end{equation}
\begin{equation}
e_{\theta,i}(t_p)\triangleq
\mathrm{wrap}\!\big(\theta_i^{\mathrm{des}}(t_p)-\hat{\theta}_i(t_p)\big)\in[-\pi,\pi].
\label{rotation}
\end{equation}
Eq.\eqref{heading} computes a desired heading angle that points in the direction of the desired planar velocity, while Eq.\eqref{rotation} compares the desired heading to the currently known heading estimate, and the $\mathrm{wrap}$ function ensures the shortest rotation direction. A proportional heading controller with gain $\gamma>0$ converts the desired velocity into translational and rotational commands:
\begin{equation}
\begin{aligned}  
\nu_i(t_p)&\triangleq \|\mathbf{v}_i^{\mathrm{des}}(t_p)\|_2\max(\cos(e_{\theta,i}(t_p)),0),\\
 \omega_i(t_p)&\triangleq\gamma\,e_{\theta,i}(t_p).
\label{eq:nu_omega}
\end{aligned}
\end{equation}
The applied input $\mathbf{u}_i(t_p) = \mathrm{sat}_\infty(\tilde{\mathbf{u}}_i(t_p), \mathbf{u}^{\max})$ is finally subject to componentwise velocity saturation with caps $\mathbf{u}^{\max} \triangleq [v^{\max}, \omega^{\max}]^\top$ (like \cite{roy2026qoc_codesign}) and saturation law:
\begin{equation}
\mathrm{sat}_\infty(\mathbf{u},\mathbf{u}^{\max}) \triangleq
\begin{bmatrix}
\max(0,\,\min(\nu,\,\nu^{\max}))\\
\max(-\omega^{\max},\,\min(\omega,\,\omega^{\max}))
\end{bmatrix}
\end{equation}

\subsection{Navigation}
\label{navigation}
Building on the simulation system model in Section \ref{dynamics}, we adopt the Nav2 and model predictive path integral (MPPI) stack \cite{mppi_main} as a more general representative candidate for real experimental validation. While the preceding framework can generalize to various planning and control algorithms, this setup utilizes a hierarchical architecture: an A* global planner \cite{a*_ref} for path generation and a rolling-horizon MPPI local controller for tracking. Both components rely on a multi-layer costmap \cite{macenski2020marathon2} (Fig. \ref{costmap}) with a static layer shown in black for known, time-invariant structures of the environment (e.g., walls and permanent obstacles) derived from a prebuilt map; a light blue dynamic layer for obstacles detected online with a LiDAR; and lastly, an inflation layer in pink that expands obstacle boundaries for extra safety using the robot footprint, localization uncertainty, and network-induced control delays. 
Formally, the workspace is discretized into a 2D grid, with $j$ denoting the direction in the x-dimension and $k$ in the y-dimension. Each cell $jk$ stores an aggregated cost $c_{jk} \in \mathbb{R}_{\ge 0}$: 
\begin{equation} c_{jk} = c^{\mathrm{stat}}_{jk} + c^{\mathrm{obs}}_{jk} + c^{\mathrm{infl}}_{jk}, \end{equation} 
with the superscripts denoting the static, obstacle, and inflation layer costs, respectively. The global planner treats this grid as a graph, computing paths that minimize cumulative cost. Similarly, MPPI evaluates candidate rollouts over horizon $\mathcal{H}$ by accumulating state-dependent costs:
\begin{equation} J_{\mathrm{map}} = \sum_{t\in\mathcal{H}} c\big(\mathbf{x}(t)\big). \end{equation} 

\begin{comment}

Formally, the 2D grid is defined as:
\begin{equation}
\mathcal{G} = \{c_{jk}\}, \qquad c_{jk}\in\mathbb{R}_{\ge 0},
\end{equation}
where each cell $c_{jk}$ stores a nonnegative traversal cost resulting from the weighted superposition of all layers. Let
\begin{equation}
c_{jk} = c^{\mathrm{stat}}_{jk} + c^{\mathrm{obs}}_{jk} + c^{\mathrm{infl}}_{jk},
\end{equation}
denote the aggregated cost at cell $(j,k)$, where the individual terms correspond to static, obstacle and inflation contributions, respectively. The global planner interprets this grid as a weighted graph and computes paths that minimize the cumulative cost over traversed cells. Similarly, MPPI incorporates the costmap into its rollout evaluation by accumulating state-dependent costs along each candidate trajectory,
\begin{equation}
J_{\mathrm{map}} = \sum_{t\in\mathcal{H}} c\big(\mathbf{x}(t)\big),
\end{equation}
where $c(\mathbf{x})$ denotes the costmap value at the cell corresponding to position $\mathbf{x}$. This unified cost representation allows spatial risk, obstacle proximity and safety margins to be directly reflected in both planning and control, while enabling online modulation of selected layers. Note that originally there are two costmaps, a local and a global one, to be used by the controller and the planner, respectively. However, for the rest of this work we assume that both costmaps are the same.
\end{comment}

This unified representation facilitates the integration of spatial risk and safety margins across both planning and control layers. In contrast to conventional Nav2 architectures, which typically employ distinct local and global costmaps, we adopt a single, shared map representation. While the current study does not exhaustively analyze these costmaps, this consolidated structure provides a formal framework for future research into network-aware navigation.

\subsection{Communication model}
\label{communication_model}
Each control/navigation update described in Section \ref{dynamics} and \ref{navigation} traverses a 5G network with a stochastic loop time $T_{\mathrm{net},i}$, comprising uplink ($T_{\mathrm{UL},i}$), compute ($T_{\mathrm{C},i}$), and downlink ($T_{\mathrm{DL},i}$) stages, as described here:
\begin{equation}
T_{\mathrm{net},i} \triangleq T_{\mathrm{UL},i} + T_{\mathrm{C},i} + T_{\mathrm{DL},i}.
\end{equation}
For a deadline $\Delta$, we define the loop completion probability:
\begin{equation}
p^{\mathrm{net}}_i(\Delta) \triangleq \Prb\big(T_{\mathrm{net},i}\le \Delta\big)=F_{\mathrm{net},i}(\Delta),
\label{p_net}
\end{equation}
where $F_{\mathrm{net},i}$ is obtained from measured/simulated compute distributions (described as compute-time model in Table \ref{tab:sim_params}). If the loop completes in time, the updated control applies to both axes; otherwise, the input stays equal to the previous sample:
\begin{equation}
\mathbf{u}_i(t_p)=
\begin{cases}
\mathbf{u}^{\mathrm{new}}_i(t_p), & T_{\mathrm{net},i}\le \Delta\\
\mathbf{u}_i(t_{p-1}), & \text{otherwise.}
\end{cases}
\label{loop_completion}
\end{equation}
This implicitly affects QoC computation discussed in Section~\ref{QOC_Model}.

\subsection{Quality of Control}
\label{QOC_Model}
To quantify the QoC for the navigation task, we define a nominal, time-varying reference trajectory $\bar{\mathbf{p}}(t)$. This reference trajectory represents the unconstrained optimal path to the destination, computed via Eq.\eqref{eq:vdes_goal_rep} by neglecting obstacles. The control performance is then characterized by the tracking error or disagreement $\boldsymbol{\delta}_i(t)$, representing the state deviation from this nominal trajectory: 
\begin{equation} 
\boldsymbol{\delta}_i(t) \triangleq \mathbf{p}_i(t)-\bar{\mathbf{p}}(t). 
\end{equation}
We define QoC from this disagreement energy. Let $W=\diag(w_x,w_y)\succ0$ be an optional axis-weighting matrix. Define the cost for robot $i$ over horizon $[0,T]$:
\begin{equation}
J^C_i \triangleq \int_0^{T}\!\|\boldsymbol{\delta}_i(t)\|_{W}^{2}\,dt,
\qquad
\|\boldsymbol{\delta}_i\|_{W}^{2}=\boldsymbol{\delta}_i^\top W\boldsymbol{\delta}_i. 
\label{AUC definition}
\end{equation}
The normalized cost or the area under the disagreement curve (AUC), denoted by $J^{C}_{\mathrm{norm},i}$, is defined as:
\begin{equation}       
J^{C}_{\mathrm{norm},i} \triangleq \int_0^{T}\!\frac{\|\boldsymbol{\delta}_i(t)\|_{W}^{2}}{\|\boldsymbol{\delta}_i(0)\|_{W}^{2}} \, dt.
\end{equation}
We map this to a bounded QoC, defined by $Q_i$, using:
\begin{equation} 
Q_i \triangleq \frac{1}{1 + \frac{\min(J^{C}_{\mathrm{norm},\max},\;  J^{C}_{\mathrm{norm},i})}{J^{C}_{\mathrm{med}}}}, \qquad Q_i \in [0,1], 
\label{eq:qi_def} 
\end{equation}
where $J^{C}_{\mathrm{norm},\max}$ is a scenario-dependent reference maximum ensuring convergence across experiments and $J^{C}_{\mathrm{med}}$ is the median AUC value. Since $\boldsymbol{\delta}_i(t)$ depends on the applied saturated control and on missed updates through ZOH, $Q_i$ captures the effects of saturation $\mathbf{u}^{\max}$, delay, and reliability.
Please note that this QoC measure is an energy-relevant proxy rather than a direct battery measurement. Larger disagreement, detours, oscillatory corrections, and stale-control recovery typically increase motion and control effort, however, direct validation against battery usage is left for future work.

\section{QoC Abstraction for AGV motion}
\label{sim_data}
\begin{figure}[htp]
  \begin{subfigure}{0.5\linewidth}
    \centering
    \includegraphics[width=1\linewidth]{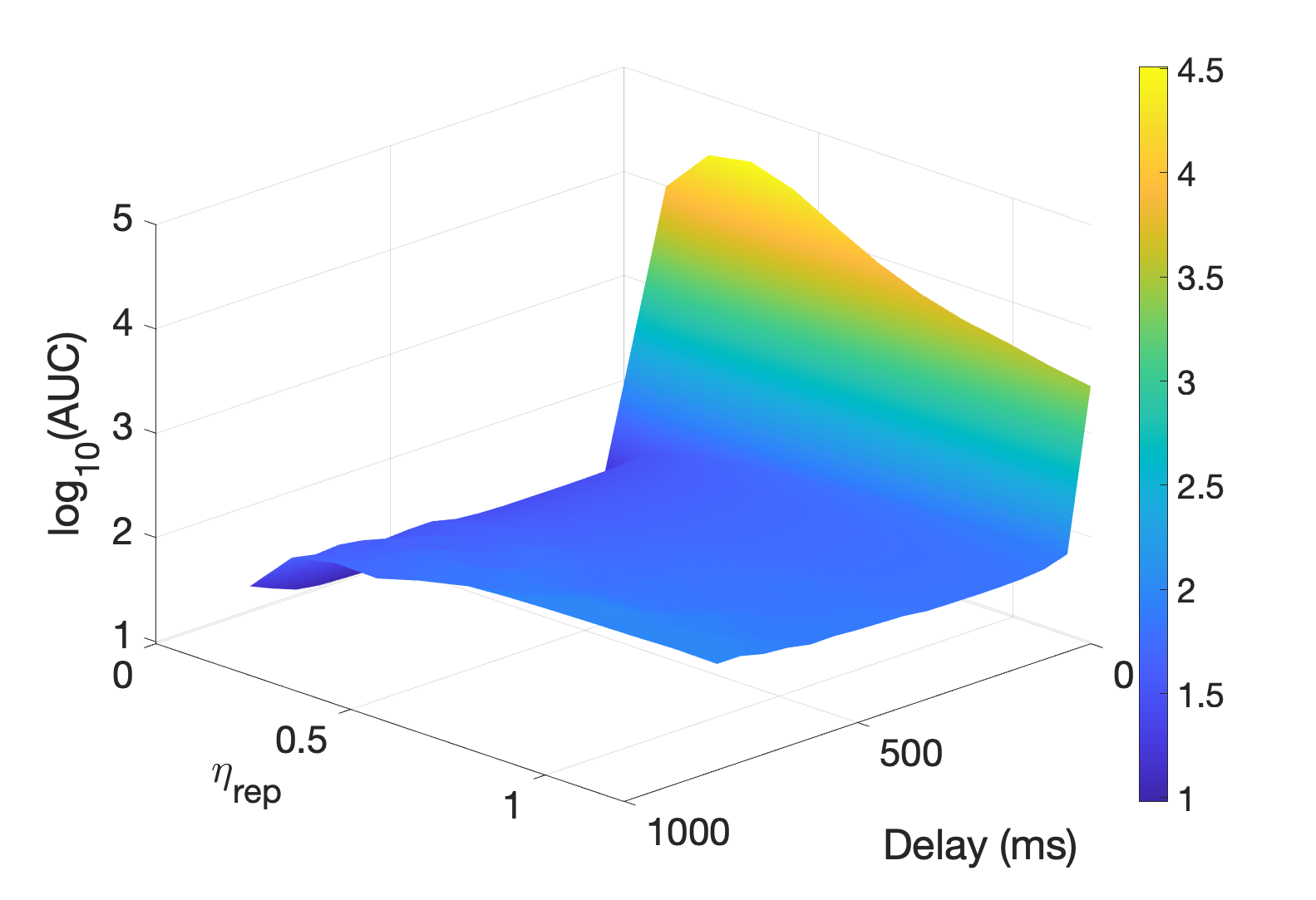}
    \caption{Repulsive Gain.}
    \label{fig:const_a}
  \end{subfigure}
  \begin{subfigure}{0.48\linewidth}
    \centering
    \includegraphics[width=1\linewidth]{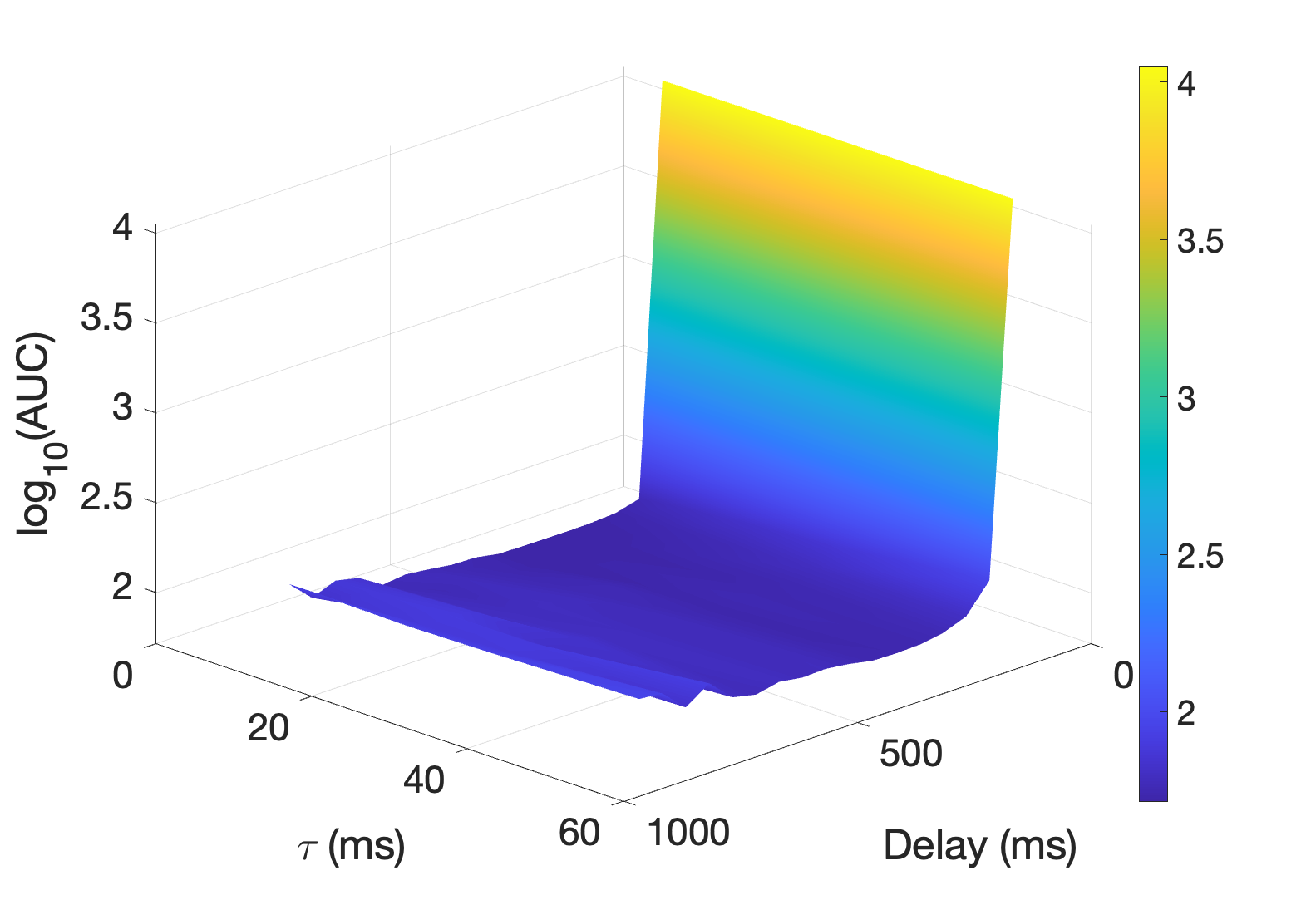}
    \caption{Sampling period.}
    \label{fig:const_b}
  \end{subfigure}
  \caption{AUC variation for selected control parameters}
  \label{fig:auc_var}
\end{figure}

We begin this section by detailing how a QoC model could be obtained, given a system of $N$ AGVs. 
The objective is to observe how Eq.\eqref{AUC definition} can help us quantify system performance with respect to various control and communication parameters together.
To realize this, we simulate the offloaded control loop where each robot evolves according to the AGV model
$\dot{\mathbf{x}}_i(t)=J(\theta_i(t))\,\mathbf{u}_i(t)$ with $\mathbf{u}_i(t)=[\nu_i(t)\;\omega_i(t)]^\top$ and ZOH updates at $t_p=t_0+p\tau$, with the parameters described in Table~\ref{tab:sim_params}. Please note that $N=2$ was simulated to match the experimental setup and the abstraction can readily scale for a higher number of robots \cite{roy2026qoc_codesign}.  For larger fleets, the aggregate network demand scales with the number of periodic state and control command streams, while the edge compute component may increase due to multi-robot planning, costmap updates, collision checking and queueing effects. Thus, scaling to multiple AGVs would primarily alter the empirical distribution of $T_{\mathrm{net},i}$ in Eq.\eqref{p_net} and the required network resource provisioning, which is precisely the type of shift captured by the QoC abstraction through $p_i^{\mathrm{net}}(\Delta)$.
Fig. \ref{fig:auc_var} shows how AUC ($J^{C}_{\mathrm{norm},i}$) varies versus delay-reliability together for selected parameters $\eta$ and $\tau$. 
Here, we can observe that at very low latency-reliability levels, AUC and thus the system performance and energy expenditure worsens substantially, across the entire range of simulated values for both the repulsive gain and the sampling period. This is attributed to very high packet drops that did not let the control commands be updated. Once sufficient reliability is achieved, AUC remains largely unchanged with different sampling periods. On the contrary, we still see worsening AUC as the repulsive gain grows together with high delays.
 %An extensive grid search over such different parameters leads us to our choice of parameters in Table~\ref{tab:sim_params} to best match the experimental setup.
\begin{table}[t]
\centering
\caption{Simulation parameters}
\label{tab:sim_params}
\renewcommand{\arraystretch}{1.08}
\setlength{\tabcolsep}{6pt}
\small
\begin{tabular}{l l}
\hline
\textbf{Parameter} & \textbf{Value} \\
\hline
Number of robots & $N=2$ \\
Simulation horizon & $T_{\mathrm{sim}}=30~\mathrm{s}$ \\
Sampling period (ZOH) & $\tau=10~\mathrm{ms}$ \\
%Integration step & $h=10^{-4}~\mathrm{s}$ \\
Compute-time model & TruncNormal$(\mu_c,\sigma_c)$ on $[0,\infty)$ \\
Compute-time mean/std & $\mu_c=0.20\mathrm{s}$,\;\;$\sigma_c=0.05\mathrm{s}$ \\
Goal and Repulsion gain & $\kappa=2$, $\eta=0.6$ \\
Speed caps & $\nu^{\max}=0.7$, $\omega^{\max}=1.4$ \\
\hline
\end{tabular}
\end{table}

\begin{comment}

\begin{figure}[t]
  \centering
  \begin{subfigure}[t]{0.49\linewidth}
    \centering
    \includegraphics[width=\linewidth]{vel_rep.eps}
    \caption{Repulsive Gain variation}
    \label{fig:vel_rep}
  \end{subfigure}\hfill
  \begin{subfigure}[t]{0.49\linewidth}
    \centering
    \includegraphics[width=\linewidth]{kappa.eps}
    \caption{Consensus gain variation}
    \label{fig:cons_gain}
  \end{subfigure}
  \caption{AUC variation vs.\ delay and different control parameters}
  \label{fig:control_var}
\end{figure}
\end{comment}

\section{Experimental Validation}
In this section, we describe how we perform experiments with the system model in Section \ref{system_model} and then perform a limited validation of the abstraction in Section \ref{sim_data}.
\subsection{Testbed and Experimental parameters}
\begin{figure}[t]
    \centering
    \includegraphics[width=0.99\linewidth]{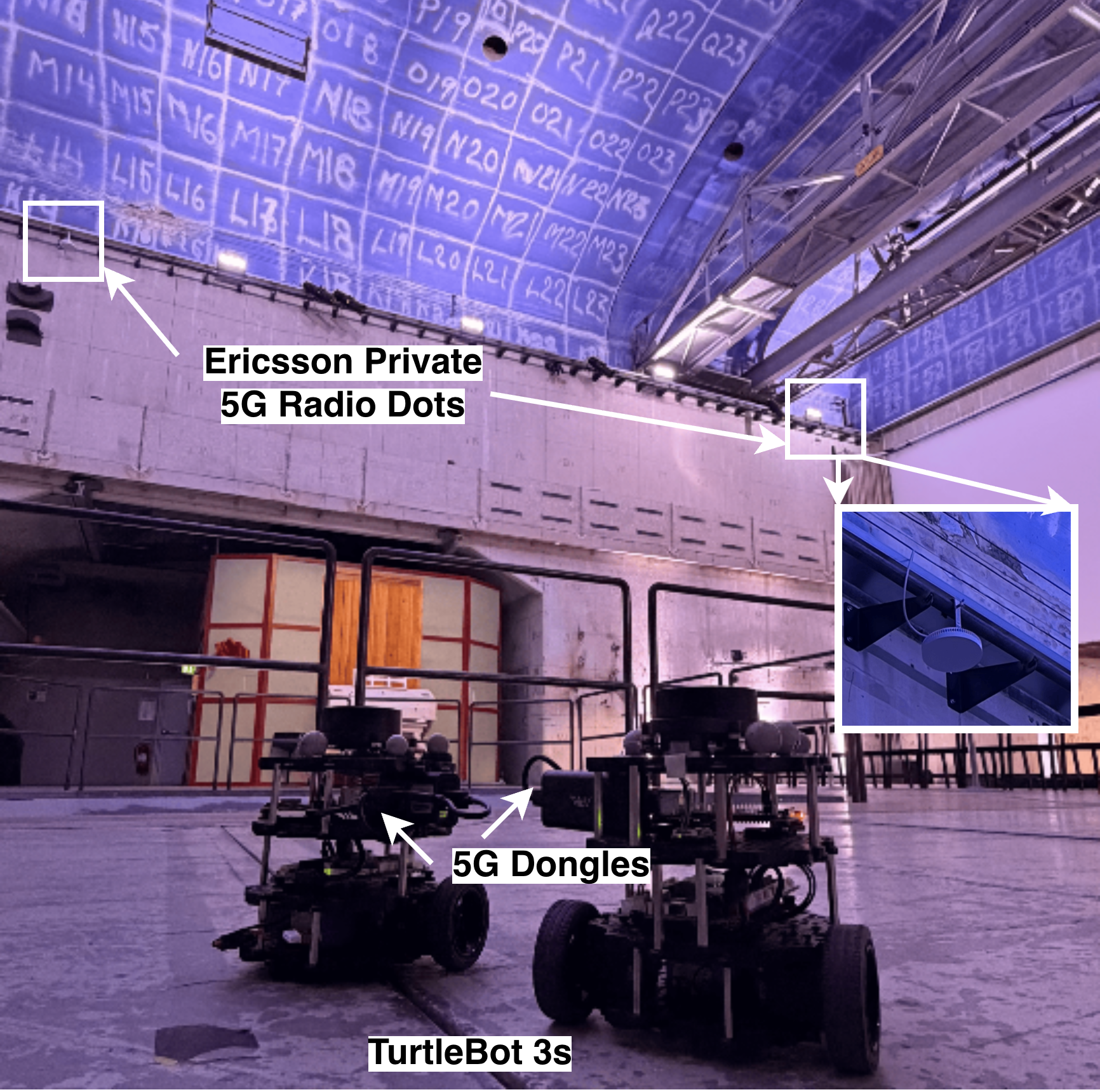}
    \caption{Private 5G testbed with TurtleBots and 5G dongles.}\label{fig:r1}
\end{figure}

Our experiments use $N=2$ TurtleBots (due to spatial, hardware and ROS 2 data distribution service (DDS) scaling constraints \cite{ros_rel}) operating on a private 5G testbed in the KTH R1 hall, with edge computation \cite{expeca}. The navigation stack for these robots runs on the edge system, and all communication occurs over the 5G network, operating on band n78. The TurtleBots were connected to the private 5G network through an external 5G dongle attached to the robot-side compute/network interface as shown in Fig. \ref{fig:r1}.

\begin{figure}[t]
    \centering
    \includegraphics[width=0.99\linewidth]{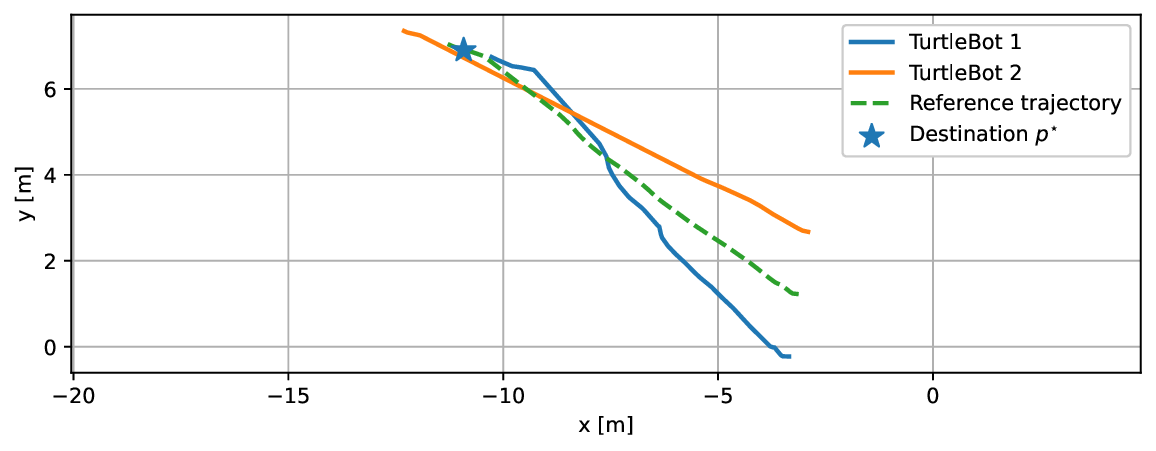}
    \caption{Adopted trajectory for a representative run of two Turtlebots performing navigation with 5G-Edge Controllers.}\label{fig:tb_trajectory}
\end{figure}

Fig.~\ref {fig:tb_trajectory} denotes how the robots are tasked with a navigation objective with dynamic collision avoidance towards destination coordinate $p^\star \in \mathbb{R}^2$.
%, which also plots the group centroid $\bar{\mathbf{p}}(t)$.
For each run, we log: (i) commanded velocities, (ii) ROS~2 QoS settings, and (iii) timestamps at UL send, edge receive, edge send, and DL receive for all control-relevant messages. To evaluate performance, 
$\mathbf{p}_i(t)$ takes the pose estimate for robot $i\in\{1,\dots,N\}$, from the robot's adaptive Monte Carlo localisation (AMCL) pose topic. We treat the AMCL pose as the best estimate of robot position in the global frame for computing disagreement-based metrics.
\begin{comment}

\begin{figure}[t]
    \centering
    \includegraphics[width=0.99\linewidth]{R1.jpg}
    \caption{The experimental private 5G testbed site at KTH R1.}\label{fig:overview}
\end{figure}
\end{comment}

%The destination $p^\star$ can be robustly estimated from the terminal portion of the experimental runs.
\subsection{Validation methodology}
\label{sec:delay_injection}
To emulate various delay-reliability combinations towards validating the abstraction simulations in Section \ref{sim_data}, we inject an artificial delay until $D_{\mathrm{cfg}}$(ms) into the velocity command stream, generated by the edge. This utilizes robot-local timestamping, eliminating the need for clock synchronization between the robot and the edge server:

\begin{itemize}
    \item At time $t_s$ on the robot clock, a source state message generated on the robot is transmitted into the navigation pipeline (where robot $i's$ AMCL pose is computed).
    \item The edge compute-based navigation stack produces a velocity command corresponding to that state.
    \item A delay node buffers the command totalling a configured duration $D_{\mathrm{cfg}}$ and only then publishes the delayed command that is actually applied by the robot base (Eq.\eqref{loop_completion}).
\end{itemize}
\subsection{Experimental Validation}
In this subsection we discuss the assumptions made and the simulation settings based on the description in Section \ref{system_model} to match the subsequent experimental settings. 
%Configured delay values $D_{\mathrm{cfg}}$ are emulated by delaying the application of the control commands computed from stale state estimates until $D_{\mathrm{cfg}}$, and by holding the previous command when an information-loop update is unavailable, as dictated by Eq. \eqref{loop_completion} and a calibrated edge compute-time distribution (truncated normal on $[0,\infty)$).
%We additionally propagate the \emph{source stamp} through the pipeline: the outgoing state message is stamped with the robot clock time at generation, and this stamp is carried forward and inserted into the header stamp of the returned \texttt{cmd\_vel} message that is ultimately applied. Thus, each applied command carries, in its header, the timestamp of the originating state sample.

%\subsection{Net Information-Loop Delay Measurement }
%\label{sec:delay_measurement}
Let $t_{s,i}^{(k)}$ be the generation time of the $k^{th}$ source state sample for robot $i$ (robot clock), i.e., the timestamp embedded in the source message header:
$t_{s,i}^{(k)} \;\triangleq\; \texttt{stamp}\big(\mathbf{x}_i(t)^{(k)}\big).$
Let $t_{a,i}^{(k)}$ be the time at which the corresponding delayed command is applied on the robot:
$t_{a,i}^{(k)} \;\triangleq\; \texttt{record}\big(\mathbf{u}_i^{(k)}\big)$,
and we recover $t_{s,i}^{(k)}$ from the propagated header stamp of the delayed command.
The configured delay is then defined as: $D_{cfg}^{(k)} \;\triangleq\; t_{a,i}^{(k)} - t_{s,i}^{(k)}.$ Provided the configured delays, and the simulation settings from Section \ref{sim_data}, let $S(\varphi)$ denote the simulated AUC at configured delay $\varphi\in\mathcal{D}$ and let
$E(\varphi)$ denote the experimentally measured AUC at the same $\varphi$. For comparison between these two metrics, we first normalize them using Eq.\eqref{eqn:auc_norm} and repeat the equivalent for $E(\varphi)$:
\begin{equation}
S(\varphi)\;=\;100\cdot\frac{S(\varphi)-\min_{\varphi'\in\mathcal{D}} S(\varphi')}
{\max_{\varphi'\in\mathcal{D}} S(\varphi')-\min_{\varphi'\in\mathcal{D}} S(\varphi')},
\label{eqn:auc_norm}
\end{equation}

\begin{comment}
    
\begin{figure}[t]
    \centering
    \includegraphics[width=0.99\linewidth]{auc_delay.eps}
    \caption{\textcolor{blue}{Simulation-based and experimental observations on the AUC variation vs delays }}\label{fig:auc_delay}
\end{figure}

\end{comment}

\begin{figure}[t]
    \centering
    \includegraphics[width=0.99\linewidth]{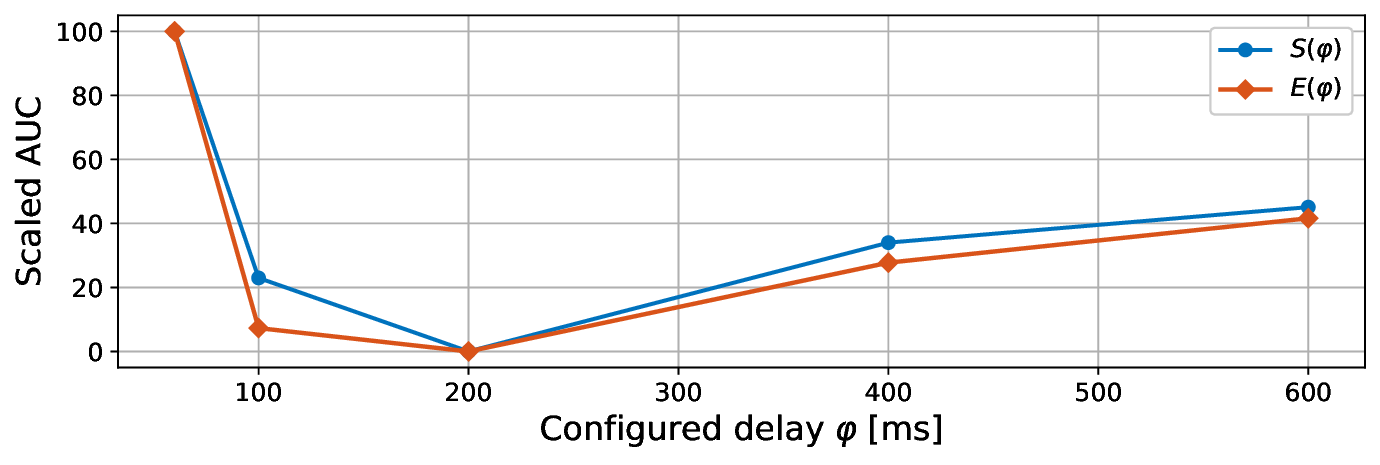}
    \caption{Normalized simulation-based and experimental observations on the AUC variation versus delays. }\label{fig:auc_delay_norm}
\end{figure}
Fig. \ref{fig:auc_delay_norm} plots how the experimental and the simulated AUC values (as in Eq.\eqref{eqn:auc_norm}) vary across different delay-reliability pairs. 
Both cases follow similar trends, consistent with findings in \cite{qoc_initial}. This alignment is reinforced further by how the reliability-vs-delay transition happens in both simulation and experiment at similar delays (e.g., reliability improving strongly around 100–200 ms), and the delay dominance region starts at similar delays as well. %At some points the values still vary since the experiment involves environment-dependent MPPI behavior. Nav2/MPPI reacts to obstacles, costmap inflation, and local minima. The simulation models collision avoidance with a simple repulsion threshold, which does not reproduce MPPI’s nuanced path choices, leading to variations.
The observed mismatch between simulation and experiments should be interpreted in light of the different roles of the collision-avoidance mechanism in the two cases. In simulation, the repulsion gain $\eta_{\mathrm{rep}}$ in Eq.\eqref{eq:repulsion} provides a compact surrogate for inter-robot collision avoidance, allowing us to isolate how delay, reliability, and control parameters affect QoC. In the experimental Nav2/MPPI stack, however, collision avoidance is not governed by a single scalar gain, and instead emerges from the interaction of obstacle-layer updates, inflation costs, MPPI rollout costs, and local path-selection effects. Hence, varying $\eta_{\mathrm{rep}}$ experimentally would require retuning several Nav2/MPPI and costmap parameters rather than changing a single physical parameter. The comparison in Fig.~\ref{fig:auc_delay_norm} is therefore intended as a trend-level validation of the QoC abstraction, showing that the low-reliability, intermediate-reliability, and delay-dominant regimes are reproduced, while exact AUC matching is limited by the richer environment-dependent behavior of the real navigation stack.

\section{Evaluations}
In this section, we first describe how the QoC abstraction simulations reveal the behavior of different parameters for AGV motion and indicate the optimal operating regimes across different parameters and then anlayze QoC experimenatlly for ROS-based reliability QoS settings.

\begin{figure}[t]
    \centering
    \includegraphics[width=0.99\linewidth]{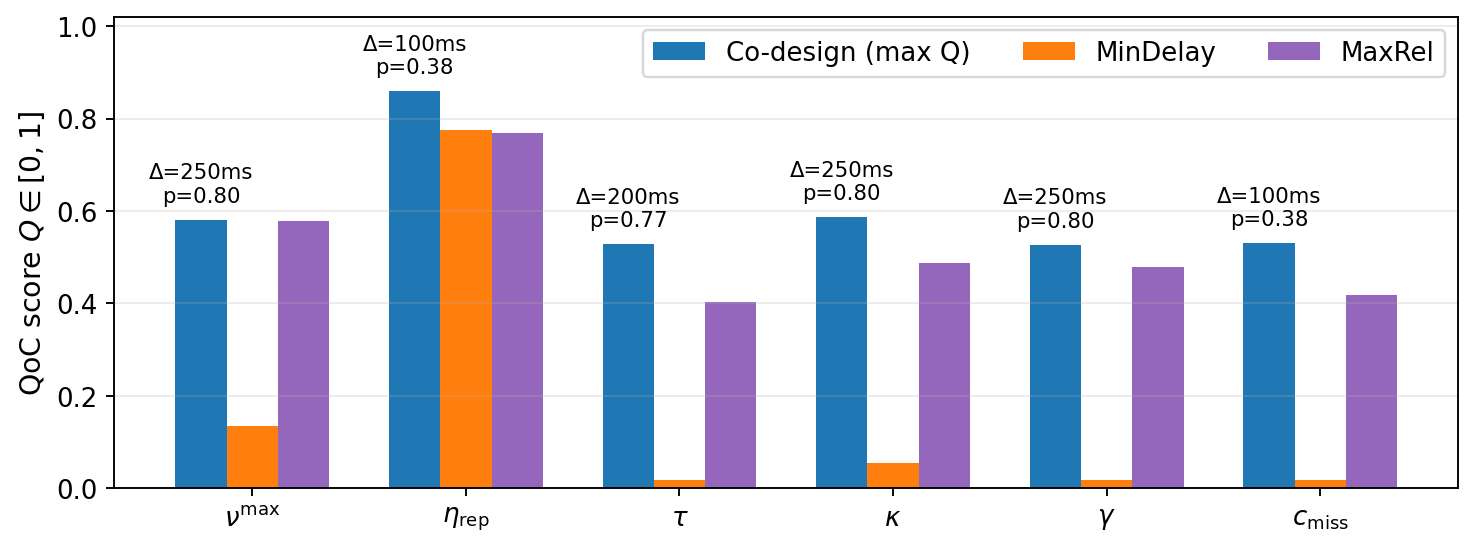}
    \caption{Parameter sweep across delay-reliability regions}\label{fig:param_sweep}
\end{figure}

\subsection{Impact of various parameters on QoC-Simulations}

In this subsection, we explore how varying different parameters crucial for AGV motion across various delay-reliability regions impacts QoC, via simulations. We further indicate optimal operating regimes towards co-design aimed at maximizing system performance and minimizing energy and state expenditure.
\label{sec:va_regimes}
For each design parameter $\Theta\in\{\nu^{\max},\eta_{\mathrm{rep}},\tau,\kappa,\gamma,c_{\mathrm{miss}}\}$ we perform a one-at-a-time sweep over a fixed grid $\Lambda_\Theta$ while keeping all other parameters at their baseline values. For every pair $(\Theta,\Delta)$ on the delay axis $\Delta\in[0,800]\,$ms, assuming enough network resources or 5G physical resource block (PRBs) are available, we evaluate $J^C_{\mathrm{norm}}$ and compute QoC.
We compare three policies per sweep: (i) \emph{Co-design} selects optimal pair $(\Theta^\star,\Delta^\star)=\arg\max_{\Theta\in\Lambda_{\Theta,\Delta}} Q(\Theta,\Delta)$; (ii) \emph{MinDelay} fixes $\Delta=\Delta_{\min}$ and tunes $\Theta$ only for that regime, $\Theta_D^\star=\arg\max_{\Theta\in\Lambda_\Theta} Q(\Theta,\Delta_{\min})$; and (iii) \emph{MaxRel} fixes $\Delta=\Delta_{\max}$ (highest loop completion probability) and tunes $\Theta$ only for that regime, $\Theta_R^\star=\arg\max_{\Theta\in\Lambda_\Theta} Q(\Theta,\Delta_{\max})$. The grouped bar chart in Fig. \ref{fig:param_sweep} reports the achieved QoC at the chosen operating point per sweep. 
It shows MinDelay can be suboptimal because low delay does not guarantee timely/reliable loop completion, leading to stale/missed updates (ZOH) and increased disagreement; MaxRel improves delivery probability but incurs large delays, making the closed-loop response sluggish. In contrast, the co-design policy selects intermediate delay-reliability regimes and corresponding parameter values that maximize QoC, consistently, highlighting the need for control--communication co-design rather than optimizing delay or reliability in isolation.
\smallskip
\subsubsection{Parameter-specific observations}
(i) $\nu^{\max}$ governs how aggressively robots can react to goal and avoidance commands; under MinDelay (low reliability), higher $\nu^{\max}$ amplifies the effect of stale headings and can increase disagreement, whereas co-design prefers moderate delays where updates are reliable enough to safely exploit higher mobility.
(ii) $\eta_{\mathrm{rep}}$ trades cohesion against safety: larger repulsion spreads robots apart and can increase disagreement, so the QoC-optimal setting tends to be moderate (or lower), while still preventing close approaches.
(iii) $\tau$ changes both the control update rate and the per-attempt transmission granularity; when reliability is low, frequent updates do not help because many are missed (ZOH dominates), but at reliable operating points an appropriate $\tau$ reduces staleness without inducing overly sluggish behavior.
(iv) $\kappa$ sets the strength of goal attraction; too large $\kappa$ under stale information can cause overshoot/oscillation (worse QoC), while too small $\kappa$ yields slow convergence, so co-design selects $\kappa$ jointly with an intermediate delay where the loop is both responsive and reliable.
(v) $\gamma$ controls heading correction; aggressive heading gains are sensitive to delayed direction estimates, so co-design improves QoC by operating in regimes where heading errors are updated reliably and $\gamma$ can be increased without instability.
(vi) $c_{\mathrm{miss}}$ inflates safety margins with age-of-information; larger values are beneficial under packet drops (more conservative motion), but become unnecessarily restrictive at high delays where the system is already sluggish, leading to a moderate co-design choice.
\subsection{ROS2 QoS-based Reliability Analysis-Experiments}
\label{sec:qos_switch}
\begin{figure}[t]
    \centering
    \includegraphics[width=0.99\linewidth]{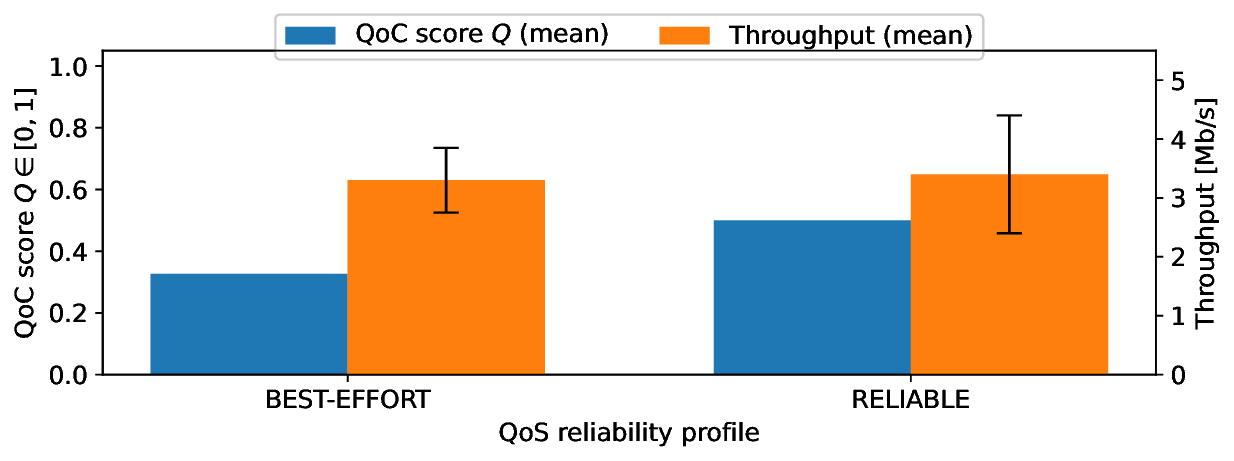}
    \caption{Experimental observations on QoC variation and throughput for different QoS reliability.}\label{fig:QoC_qos}
\end{figure}

\begin{figure}[t]
    \centering
    \includegraphics[width=0.99\linewidth]{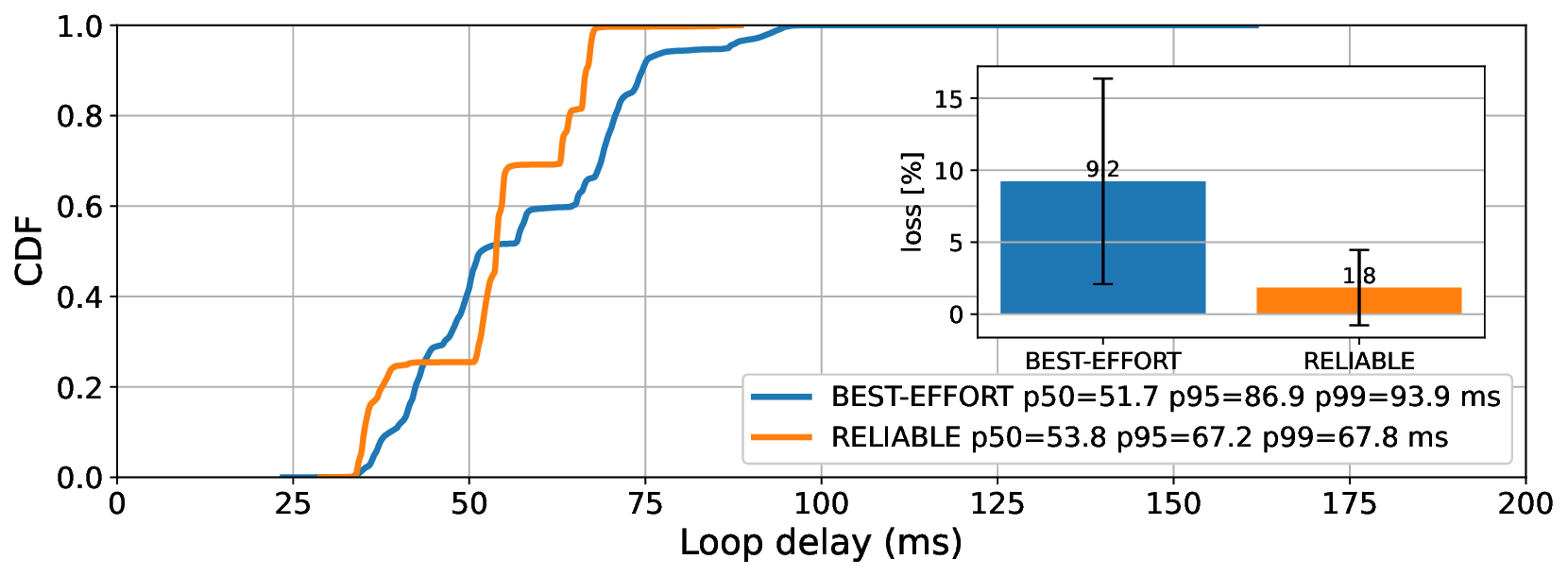}
    \caption{CDF of observed delay distributions and packet loss percentage for different QoS reliability.}\label{fig:deadline}
\end{figure}

A key parameter for practical experiments is the ROS 2 DDS middleware, which provides QoS policies to trade off delivery guarantees and latency \cite{ros_rel}. In this evaluation, we analyze how QoC varies across different QoS settings, experimentally. We focus on the reliability policy, implemented at the application layer over user datagram protocol (UDP). Within this, BEST-EFFORT minimizes overheads by preceding retransmissions, prioritizing low latency while tolerating packet loss. Conversely, RELIABLE ensures delivery through application-layer retransmissions, resulting in latency characteristics distinct from transmission control protocol (TCP).
%, at the cost of increased delay.
%\subsubsection{BEST-EFFORT vs RELIABLE:} In this evaluation, we perform further experimental runs with these two policies to better understand their implications on QoC. 
Fig. \ref{fig:QoC_qos}, plots the QoC score, which is significantly lower under BEST-EFFORT QoS, as robots frequently miss timely updates, degrading cohesion. This is further reinforced by the delay-reliability relations, leading to higher packet deadline misses under BEST EFFORT QoS, as captured by the cumulative distribution function (CDF) and packet loss quantification in Fig. \ref{fig:deadline}, which also shows the 50, 95, and 99 percentile delays for both QoSs. While RELIABLE QoS restores performance (51.5 \% higher QoC), it introduces higher throughput variability (leading to higher 5G PRB demand if resources are dimensioned) and demand, as illustrated by the increased mean and standard deviation of throughput. This provides a critical trade-off for network providers, as it implies that a system optimized for high QoC may require more aggressive resource over-provisioning at the 5G base station. These results confirm that reliability choices directly shape the measured QoC and energy efficiency.

\begin{comment}

\textcolor{red}{Show a plot with only Reliable, only Best Effort and Smart Switching based comparison}
Fig. \ref{fig:rel_val} shows how changes in reliability QoS levels lead to variations in QoS across different speed levels and costmap network conditions.

\begin{figure}[t]
    \centering
    \includegraphics[width=0.99\linewidth]{reliability.png}
    \caption{Experimental observations on QoC variation for different reliability QoS levels \textcolor{red}{Figure to be replaced with vanilla comparison of only best effort, only Reliable and Switched QoS policy}}\label{fig:rel_val}
\end{figure}

\begin{figure}[t]
    \centering
    \includegraphics[width=0.99\linewidth]{ros_rel.eps}
    \caption{Experimental observations on mean AUC variation and throughput for different QoS reliability}\label{fig:rel_val_qos}
\end{figure}
\end{comment}

\begin{comment}

\subsubsection{Reliability QoS with delay constraints:}
In this evaluation, we cover three cases within the RELIABLE Reliability QoS level. Fig. \ref{fig:QoC_qos_reliable} shows that the unconstrained reliable case achieves the highest QoC, suggesting the most consistent coordination when reliability is enforced without additionally restricting delay. Introducing explicit delay constraints reduces the QoC score, most noticeably for the 60 ms constraint. This indicates that tighter latency policing can degrade closed-loop behavior by increasing drop/hold behavior and inducing less stable command updates. The 100 ms constraint partially recovers QoC relative to 60 ms, implying that a slightly looser deadline can better balance timeliness and reliability, yielding smoother multi-robot coordination.
\begin{figure}[t]
    \centering
    \includegraphics[width=0.99\linewidth]{reliable_3case_qoc_eq20.eps}
    \caption{QoC variation for different delay constraints based QoS reliability.}\label{fig:QoC_qos_reliable}
\end{figure}
\end{comment}
\begin{comment}

\begin{figure}[t]
    \centering
    \includegraphics[width=0.99\linewidth]{throughput_AUC.png}
    \caption{Experimental observations on the throughput variation for different reliability QoS levels across multiple runs}\label{fig:thr_val_qos}
\end{figure}

\end{comment}
\section{Conclusions and Future Work}
This work introduces a validated QoC-based abstraction designed to characterize the performance of collaborative robotic systems under stochastic network conditions. By incorporating non-holonomic kinematic constraints, we extend past QoC frameworks to bridge the gap between theoretical models and practical, edge-offloaded AGV navigation with dynamic collision avoidance. Towards this, we provide optimal operating points for various parameters for different delay-reliability points. Our experimental results, gathered over a private 5G testbed, closely align with simulation-based predictions, confirming the framework's utility. Furthermore, a comparative analysis of ROS~2 transport policies demonstrates that RELIABLE QoS yields a 51.5\% improvement in QoC over BEST-EFFORT, albeit at the expense of increased throughput variability. These findings provide a principled foundation for control-communication co-design in next-generation industrial automation.
Future work will further calibrate the surrogate repulsion model against the Nav2/MPPI collision-avoidance stack by jointly varying costmap inflation, MPPI weights, and inter-robot spacing in larger experimental campaigns.

% =========================================================
\section{Acknowledgements}
This work was supported in part by (1) Digital Futures project ‘Toward Safe Smart Construction: Algorithms, Digital Twins and Infrastructures’ under Grant VF 2020-0315, in part by (2) VR (Swedish Research Council) project ‘Optimal Sampling for Interactive Networked Applications’ under Grant 2022-03922 and lastly, (3) Styffes stiftelse.

\bibliographystyle{IEEEtran}
\bibliography{ref}

@INPROCEEDINGS{ros_rel,
  author={Park, Hyung-Seok and Lee, Sanghoon and Um, Doosik and Ryu, Hyunho and Park, Kyung-Joon},
  booktitle={IEEE INFOCOM 2025 - IEEE Conference on Computer Communications}, 
  title={An Analytical Latency Model of the Data Distribution Service in ROS 2}, 
  year={2025},
  volume={},
  number={},
  pages={1-10},
  doi={10.1109/INFOCOM55648.2025.11044454}}

@inproceedings{expeca,
author = {Mostafavi, Samie and Moothedath, Vishnu Narayanan and Ronngren, Stefan and Roy, Neelabhro and Sharma, Gourav Prateek and Seo, Sangwon and Munoz, Manuel Olguin and Gross, James},
title = {ExPECA: An Experimental Platform for Trustworthy Edge Computing Applications},
year = {2024},
isbn = {9798400701238},
publisher = {Association for Computing Machinery},
address = {New York, NY, USA},
url = {https://doi.org/10.1145/3583740.3626819},
doi = {10.1145/3583740.3626819},
booktitle = {Proceedings of the Eighth ACM/IEEE Symposium on Edge Computing},
pages = {294–299},
numpages = {6},
location = {Wilmington, DE, USA},
series = {SEC '23}
}

@ARTICLE{mppi_main,
  author={Williams, Grady and Drews, Paul and Goldfain, Brian and Rehg, James M. and Theodorou, Evangelos A.},
  journal={IEEE Transactions on Robotics}, 
  title={Information-Theoretic Model Predictive Control: Theory and Applications to Autonomous Driving}, 
  year={2018},
  volume={34},
  number={6},
  pages={1603-1622},
  doi={10.1109/TRO.2018.2865891}}

@article{baxi2022towards,
  title={Towards factory-scale edge robotic systems: Challenges and research directions},
  author={Baxi, Amit and Eisen, Mark and Sudhakaran, Susruth and Oboril, Fabian and Murthy, Girish S and Mageshkumar, Vincent S and Paulitsch, Michael and Huang, Margaret},
  journal={IEEE Internet of Things Magazine},
  volume={5},
  number={3},
  pages={26--31},
  year={2022},
  publisher={IEEE}
}

@ARTICLE{codesign1,
  author={Qiao, Yue and Fu, Yusun and Yuan, Muyun},
  journal={IEEE Internet of Things Journal}, 
  title={Communication–Control Co-Design in Wireless Networks: A Cloud Control AGV Example}, 
  year={2023},
  volume={10},
  number={3},
  pages={2346-2359},
  doi={10.1109/JIOT.2022.3211766}}

@INPROCEEDINGS{mppi,
  author={Williams, Grady and Drews, Paul and Goldfain, Brian and Rehg, James M. and Theodorou, Evangelos A.},
  booktitle={2016 IEEE International Conference on Robotics and Automation (ICRA)}, 
  title={Aggressive driving with model predictive path integral control}, 
  year={2016},
  volume={},
  number={},
  pages={1433-1440},
  doi={10.1109/ICRA.2016.7487277}}

@INPROCEEDINGS{qoc_initial,
  author={Roy, Neelabhro and Dhullipalla, Mani H. and Sharma, Gourav Prateek and Dimarogonas, Dimos V. and Gross, James},
  booktitle={2025 IEEE 22nd Consumer Communications \& Networking Conference (CCNC)}, 
  title={Quality of Control Based Resource Dimensioning for Collaborative Edge Robotics}, 
  year={2025},
  volume={},
  number={},
  pages={1-7},
  doi={10.1109/CCNC54725.2025.10976180}}

@article{carabin2017review,
  title={A review on energy-saving optimization methods for robotic and automatic systems},
  author={Carabin, Giovanni and Wehrle, Erich and Vidoni, Renato},
  journal={Robotics},
  volume={6},
  number={4},
  pages={39},
  year={2017},
  publisher={MDPI}
}

@article{mohanti2023norm,
  title={L-norm: Learning and network orchestration at the edge for robot connectivity and mobility in factory floor environments},
  author={Mohanti, Subhramoy and Roy, Debashri and Eisen, Mark and Cavalcanti, Dave and Chowdhury, Kaushik},
  journal={IEEE Transactions on Mobile Computing},
  volume={23},
  number={4},
  pages={2898--2914},
  year={2023},
  publisher={IEEE}
}

@ARTICLE{ros_distributed,
  author={Urbaniak, Dominik and Bro Damsgaard, Sebastian and Zhang, Weifan and Rosell, Jan and Suárez, Raúl and Suppa, Michael},
  journal={IEEE Access}, 
  title={Distributed Control for Collaborative Robotic Systems Using 5G Edge Computing}, 
  year={2024},
  volume={12},
  number={},
  pages={148706-148718},
  doi={10.1109/ACCESS.2024.3475584}}

@inproceedings{cleave,
  author    = {Olgu\'{\i}n Mu\~{n}oz, Manuel and Roy, Neelabhro and Gross, James},
  title     = {{CLEAVE}: Scalable and Edge-Native Benchmarking of Networked Control Systems},
  year      = {2022},
  isbn      = {9781450392532},
  publisher = {Association for Computing Machinery},
  address   = {New York, NY, USA},
  doi       = {10.1145/3517206.3526272},
  booktitle = {Proceedings of the 5th International Workshop on Edge Systems, Analytics and Networking},
  pages     = {37-42},
  numpages  = {6},
  location  = {Rennes, France},
  series    = {EdgeSys '22}
}

@ARTICLE{latency_min,
  author={Sardar, Asif Ahmed and Rao, Aravinda S. and Alpcan, Tansu and Das, Goutam and Palaniswami, Marimuthu},
  journal={IEEE Transactions on Cognitive Communications and Networking}, 
  title={Network Resource Allocation for Industry 4.0 With Delay and Safety Constraints}, 
  year={2024},
  volume={10},
  number={1},
  pages={223-237},
  doi={10.1109/TCCN.2023.3319529}}

@ARTICLE{a*_ref,
AUTHOR = {Xuan, Doan Thanh and Hung, Nguyen Thanh and Thang, Vu Toan},
TITLE = {A Comprehensive Review of Improved A* Path Planning Algorithms and Their Hybrid Integrations},
JOURNAL = {Automation},
VOLUME = {6},
YEAR = {2025},
NUMBER = {4},
ARTICLE-NUMBER = {52},
URL = {https://www.mdpi.com/2673-4052/6/4/52},
ISSN = {2673-4052},
DOI = {10.3390/automation6040052}
}

@InProceedings{macenski2020marathon2,
author = {Macenski, Steven and Martin, Francisco and White, Ruffin and Ginés Clavero, Jonatan},
title = {The Marathon 2: A Navigation System},
booktitle = {2020 IEEE/RSJ International Conference on Intelligent Robots and Systems (IROS)},
year = {2020}
}

@INPROCEEDINGS{AGV_model,
  author={Tayade, Shreya and Rost, Peter and Maeder, Andreas and Schotten, Hans D.},
  booktitle={2020 IEEE International Conference on Communications Workshops (ICC Workshops)}, 
  title={Impact of Short Blocklength Coding on Stability of an AGV Control System in Industry 4.0}, 
  year={2020},
  volume={},
  number={},
  pages={1-6},
  doi={10.1109/ICCWorkshops49005.2020.9145394}}

@article{QoC_Assessment,
  title={Assessing quality of control in tactile cyber--physical systems},
  author={Polachan, Kurian and Pa, Joydeep and Singh, Chandramani and Prabhakar, TV},
  journal={IEEE Transactions on Network and Service Management},
  volume={19},
  number={4},
  pages={5348--5365},
  year={2022},
  publisher={IEEE}
}

@ARTICLE{roy2026qoc_codesign,
  author={Roy, Neelabhro and Dhullipalla, Mani H. and Sharma, Gourav Prateek and Sandberg, Sara and Dimarogonas, Dimos V. and Gross, James},
  journal={IEEE Transactions on Industrial Informatics}, 
  title={Quality of Control-Based Control-Communication Co-Design for Collaborative Robotics}, 
  year={2026},
  volume={22},
  number={6},
  pages={5219-5230},
  doi={10.1109/TII.2026.3664663}}

@INPROCEEDINGS{apf,
  author={Singletary, Andrew and Klingebiel, Karl and Bourne, Joseph and Browning, Andrew and Tokumaru, Phil and Ames, Aaron},
  booktitle={2021 IEEE/RSJ International Conference on Intelligent Robots and Systems (IROS)}, 
  title={Comparative Analysis of Control Barrier Functions and Artificial Potential Fields for Obstacle Avoidance}, 
  year={2021},
  volume={},
  number={},
  pages={8129-8136},
  doi={10.1109/IROS51168.2021.9636670}}

@ARTICLE{zhihao,
  author={Liu, Zhihao and Silva, Jorge and Zhong, Ruirui and Qin, Qiang and Roy, Neelabhro and Nan Fernandez-Ayala, Victor and Lesko, Johan and Håkansson, Ulf and Sandberg, Sara and Dimarogonas, Dimos V. and Gross, James and Vincent Wang, Xi and Wang, Lihui},
  journal={IEEE Transactions on Systems, Man, and Cybernetics: Systems}, 
  title={ConstrucTwin: Digital Twin-Driven Multirobot Construction System Toward Industry 5.0}, 
  year={2026},
  volume={},
  number={},
  pages={1-16},
  doi={10.1109/TSMC.2026.3658622}}

@ARTICLE{collab4g,
  author={Groshev, Milan and Zanzi, Lanfranco and Delgado, Carmen and Li, Xi and Oliva, Antonio de la and Costa-Pérez, Xavier},
  journal={IEEE Transactions on Network and Service Management}, 
  title={Energy-Aware Joint Orchestration of 5G and Robots: Experimental Testbed and Field Validation}, 
  year={2025},
  volume={22},
  number={4},
  pages={3046-3059},
  doi={10.1109/TNSM.2025.3555126}}

\end{document}